\documentclass[letterpaper, 10 pt, conference]{ieeeconf}

\usepackage{graphicx} 
\usepackage{xcolor} 
\usepackage{multirow}
\usepackage{tabularray}
\usepackage{cite}
\usepackage{array}
\usepackage{amsmath}
\usepackage{booktabs}
\usepackage{multirow}
\usepackage{makecell}
\usepackage[font=footnotesize,labelsep=period]{caption}
\usepackage{balance}
\usepackage{subcaption}
\usepackage{adjustbox}
\usepackage{geometry}

\IEEEoverridecommandlockouts
\begin{document}

\title{Coastal Environment Generation with HoloOcean}
\author{Abigail Austin, Brady Moon, and Joshua G. Mangelson
    \thanks{This work was funded under Department of Navy awards N00014-24-1-2503 and N00014-24-1-2301 issued by the Office of Naval Research.}
  \thanks{A.~Austin, B.~Moon, and J.~Mangelson are at Brigham Young University. They can be reached at: \texttt{\{abiausti, brady.moon, mangelson\}@byu.edu}. }
}

\makeatletter
\let\@oldmaketitle\@maketitle
\renewcommand{\@maketitle}{\@oldmaketitle
\centering
\begin{tabular}{ccc}
\includegraphics[width=.98\textwidth]{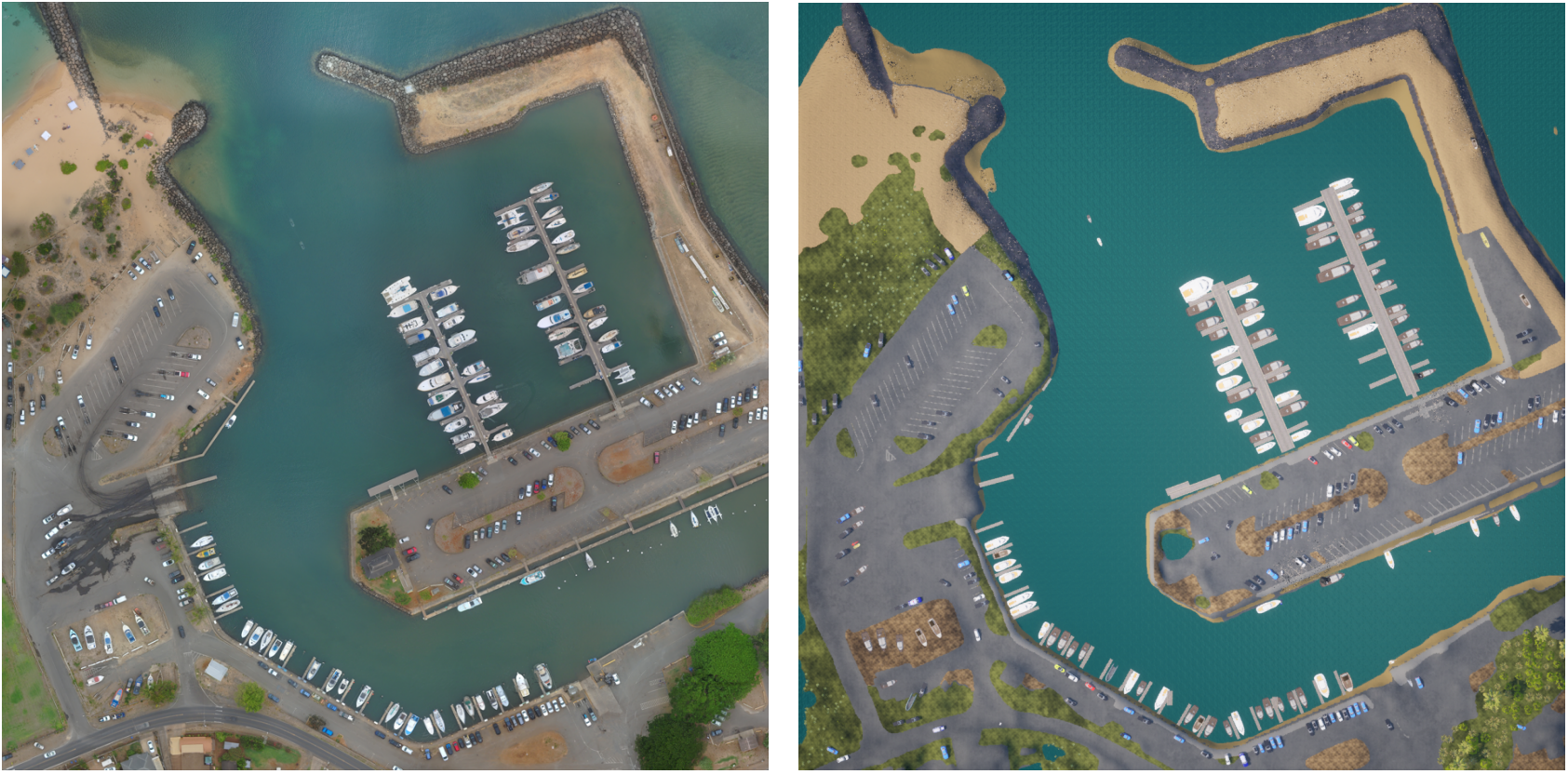}
\end{tabular}
\captionof{figure}{Qualitative comparison between the real-world harbor scene (left) and
    the corresponding replicated scene made using our pipeline (right).
}
\label{fig:real_harbor_comparison}
}
\makeatother

\maketitle
\setcounter{figure}{1}

\begin{abstract}
Marine robotic simulation provides a safe and inexpensive method of developing and testing algorithms for unmanned underwater vehicle (UUV) and unmanned surface vessel (USV) autonomy and perception before full field deployment. However, these simulations are often limited by the availability of simulated environments. Current marine robotics simulation suites offer manual ways to edit or create environments, but they require existing data or specialized knowledge of the environment system. To address these issues, we introduce a novel Unreal Engine 5 level generation pipeline that enables automatic creation of coastal environments for HoloOcean. Our pipeline relies on a user-provided overhead image of a coastal scene. The pipeline then uses the image to generate height map data, as well as automatically select assets and place them in the environment.
\end{abstract}

\section{Introduction}
\label{sec:intro}

Marine robotic simulation provides a safe and inexpensive method for developing and evaluating algorithms for autonomy and perception of unmanned underwater vehicles (UUV) and unmanned surface vessels (USV) prior to full field deployment. Recently, several marine robotics simulation suites have been developed to address this need~\cite{potokarHoloOcean2022, song2025oceansim, UNav-Sim, stonefish}. However, the fidelity of the simulation depends on the fidelity of the simulated environment, and in many cases, environment creation becomes the limiting factor to large-scale high-fidelity testing. Most simulation environments provide a manual method for modifying or creating novel environments; however, creating such environments requires significant time, data, and expert knowledge.

\begin{figure*}[t]
    \centering
    \includegraphics[width=\linewidth]{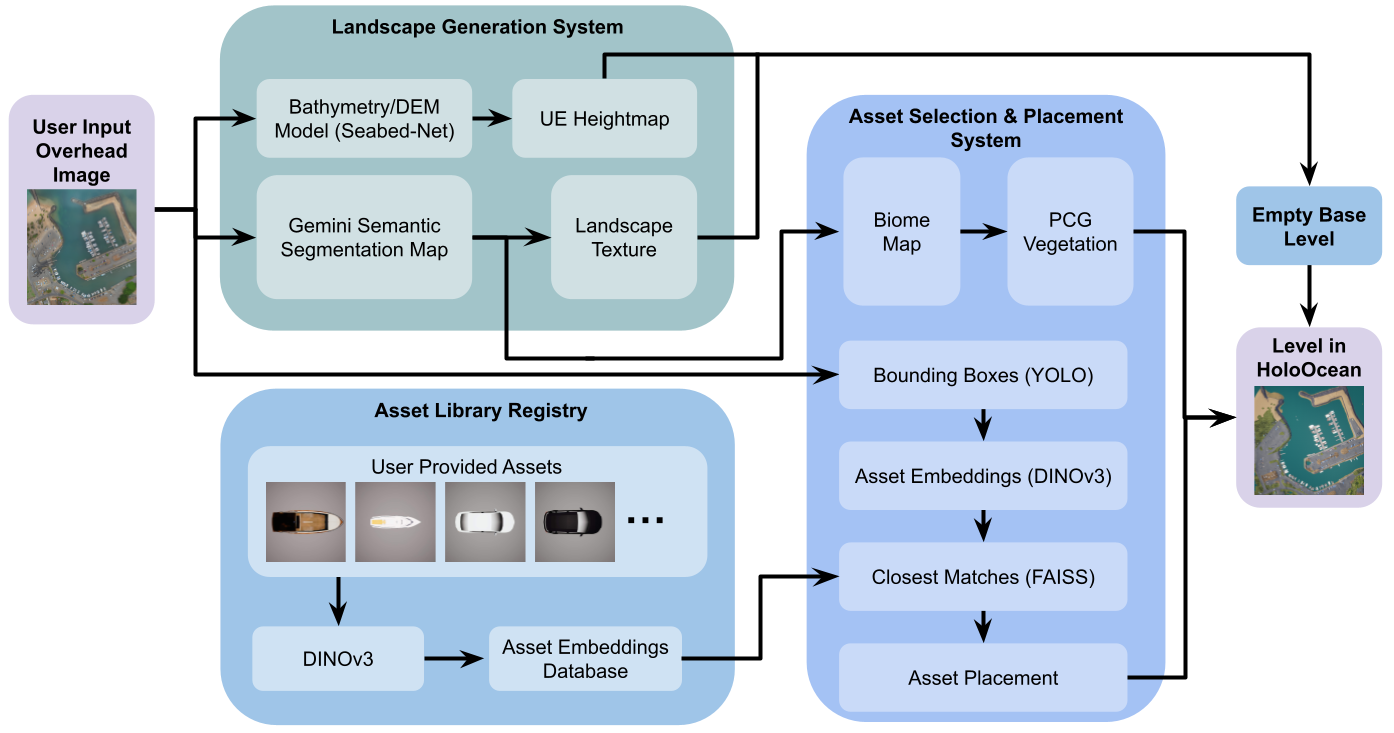}
    \caption{Flow diagram demonstrating proposed pipeline to interface with UE 5. The user-provided overhead image is input to a Landscape Generation System as well as an Asset Selection and Placement System to create a final environment for HoloOcean.}
    \label{fig:highlevel}
\end{figure*}

In this paper, we introduce a novel Unreal Engine 5 (UE)~\cite{unrealengine} level generation pipeline that enables automatic creation of coastal simulation environments for HoloOcean~\cite{potokarHoloOcean2022}, an open source marine robotics simulator suite. Our system takes an overhead image as input and automatically generates a simulated test environment based upon that image as depicted in Fig.~\ref{fig:real_harbor_comparison}. To facilitate this, we present an automatic terrain generation tool that incorporates semantic information from the overhead imagery and estimates a height map for the terrain. We then present an automated asset selection and placement system that enables incorporation of assets from an asset library. Finally, we leverage both of these techniques to build a full coastal environment generation pipeline that automatically builds an Unreal environment from a single overhead image. 

The specific contributions of this paper include:
\begin{enumerate}
    \item A novel tool-set for automatic asset selection and placement;
    \item A full image-to-environment pipeline for coastal scene generation; and
    \item An evaluation of the pipeline and techniques.
\end{enumerate}

The pipeline allows researchers to utilize HoloOcean with customized levels while removing the need for expert UE environment development knowledge. The resulting levels are fully compatible with HoloOcean's various sonar implementations and robotic agents.

\section{Related Work}

\subsection{HoloOcean}

HoloOcean is built off of UE 5, and as such is able to employ UE's level design tools, assets, and rendering pipelines, thus allowing users to build upon a major game-design user-base and community. HoloOcean offers prebuilt levels with parameters to change currents, underwater appearance, and FFT waves. UE 5 released a procedural content generation (PCG) system~\cite{unrealPCG} that enables procedurally placed assets and vegetation within a level. HoloOcean can utilize PCG to create more realistic environments. However, learning and using the UE level design tools to create custom environments for HoloOcean requires a significant amount of time and precision to achieve desired results. 

\subsection{Approaches to Marine Environment Simulation}
Marine robotics simulators have various approaches to custom user environments. Project DAVE~\cite{projectDAVE} (based on Gazebo) allows users to input bathymetry to create heightmaps, but lacks the visuals and access to assets found with UE. OceanSim~\cite{song2025oceansim} (based upon NVIDIA IsaacSim) allows users to upload 3D scans of real world locations to use in simulation, but this requires obtaining real-world models of an environment. Stonefish~\cite{stonefish} allows users to load in assets or alter parameters, such as ocean appearance, underwater currents, and wind, but is limited to a predefined level. UNav-Sim~\cite{UNav-Sim}, similar to HoloOcean, is built off of UE 5 and allows users to access UE tools for level design.

\subsection{Environment Generation}

Many current methods of environment generation rely on PCG to enhance the realism of generated environments. Infinigen~\cite{infinigen2023infinite} allows users to generate diverse scenes with config scripts in Blender through their extensive PCG template library for assets, materials, and animals. Infinigen scenes work well for computer vision tasks, as they have automatic object annotations. These scenes can be transferred from Blender to UE, but require manually reassembling the materials since Blender's materials do not easily export to Unreal. 

SceneX~\cite{zhou2024scenexproceduralcontrollablelargescale} and UnrealLLM~\cite{songtang-etal-2025-unrealllm} extend Infinigen by allowing users to input text prompts to generate scenes in Blender or UE respectively. Both integrate LLMs to control parameters for PCG templates, but neither are publicly available. AutoUE~\cite{autoue} builds on both papers by implementing a pipeline to generate both scenes and simplistic game play in UE and is publicly available. Although text prompts to these models can be fine tuned, the approach is not adequate for replicating real world locations. Other programs such as Cesium~\cite{cesium} allow users to load GPS coordinates and retrieve landscape data, but often lack details for assets and underwater spaces and require further fine tuning either with procedural content or by manually editing the environment.

\subsection{DEM and Bathymetry Generation}

Predictive models for Digital Elevation Models (DEM) and bathymetry is an ongoing challenge. Mandani \textit{et al.}~\cite{DEM_w_deeplearning} use a generative adversarial network to predict DEM from satellite imagery for mountainous regions, with limited results on lowland areas. LunarDepthNet~\cite{DEM_lunar} use a UNet-style encoder-decoder architecture to predict lunar DEMs from grayscale monocular images. Pixels2Peaks~\cite{pixels2peaks} utilizes a diffusion model with loss for different terrain components such as hydrology and geomorphology consistency to reconstruct terrain from a single image. None of these approaches account for bathymetry, and have very focused applications with either mountain or lunar regions.

Seabed-Net~\cite{seabednet} is a bathymetry and pixel classification prediction model trained on a novel coastal dataset from MagicBathyNet~\cite{magicbathynet}. Seabed-Net provides two different models trained on separate locations. The models perform well in their corresponding locations but struggle to generalize to other coastal regions.

\begin{figure}[t]
    \centering
    \includegraphics[width=\columnwidth]{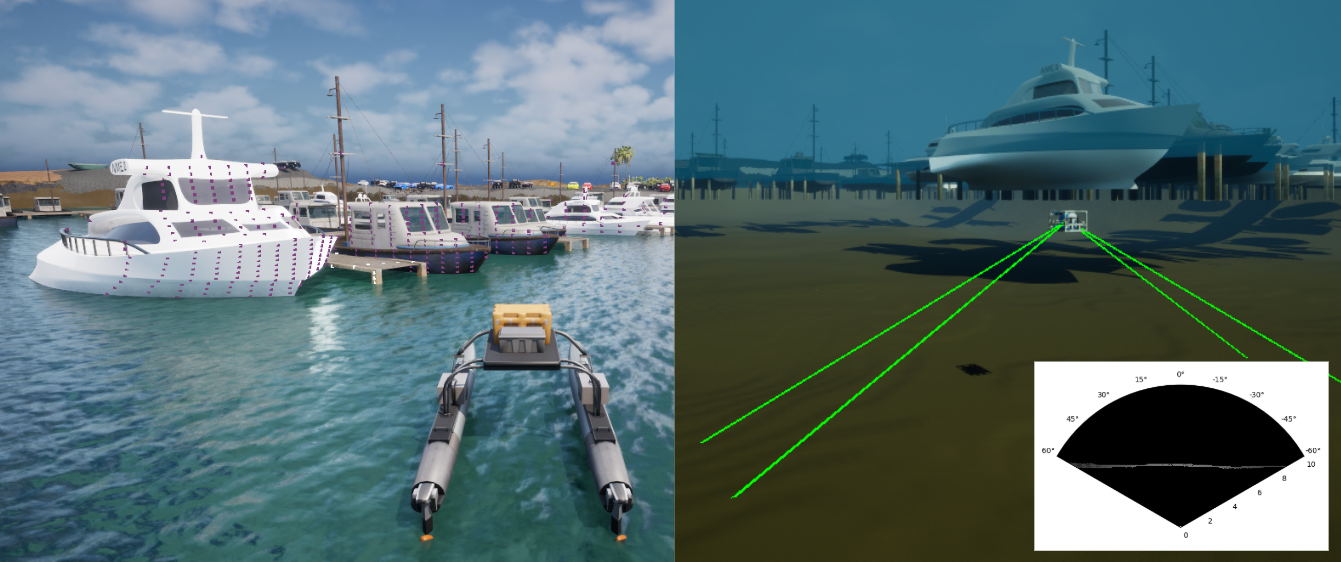}
    \caption{Generated environment with Raycast Semantic LiDAR identifying material types and Imaging Sonar results.}
    \label{fig:lidar}
\end{figure}

\section{Coastal Environment Generation Pipeline}

In this paper, we produce a coastal environment generation toolset that only requires a true (or generated) overhead image of the desired scene. The output is a fully functional HoloOcean simulation environment (or level).
There are two main components of the pipeline: a landscape generation system and an asset selection and placement system. In this section, we outline the full pipeline illustrated in Fig.~\ref{fig:highlevel}.

\subsection{Base Level Setup}

To support the pipeline, we utilize a premade mostly empty base-level. This level allows us to utilize UE editor tools and contains the HoloOcean weather manager, a water plane, correct lighting, and post process volumes for underwater appearance controls. The landscape and assets are spawned in this base level in later steps.

\subsection{Image Input and Semantic Processing}
\label{sec:Semantic}

Users input a single aerial image of a coastline that they wish to replicate within HoloOcean, along with dimensions of the image in meters. We then prompt Google Gemini~\cite{google_gemini} with the image to first remove objects such as boats and cars and then create a semantic segmentation map on that cleaned image. We prompt Gemini to label pixels with eight different terrain types: Asphalt, Water, Grass, Trees, Dirt, Rocks, Sand, and Sidewalk/Parking Lines. This semantic segmentation map is used later in the pipeline to automatically texture the landscape and as a biome map for vegetation. 

\subsection{Landscape Generation System}

For this paper, we utilized Seabed-Net to generate height maps. We apply the height map to an Unreal landscape and scale the landscape based on the input user image. A water plane is spawned at $z=0$. 

The semantic segmentation map created in step \ref{sec:Semantic} is used as a texture input to a material function that filters each pixel color to assign a material type at that location. Fig.~\ref{fig:lidar} demonstrates material type detection with the HoloOcean Raycast Semantic LiDAR sensor as well as Imaging Sonar results for the landscape. Each material type has a color multiplier that users can tune to adjust the color of that material. The material functions for each material type can also be replaced with user provided material functions to better fit their requirements.

\begin{figure}[t]
    \centering
    \includegraphics[width=\columnwidth]{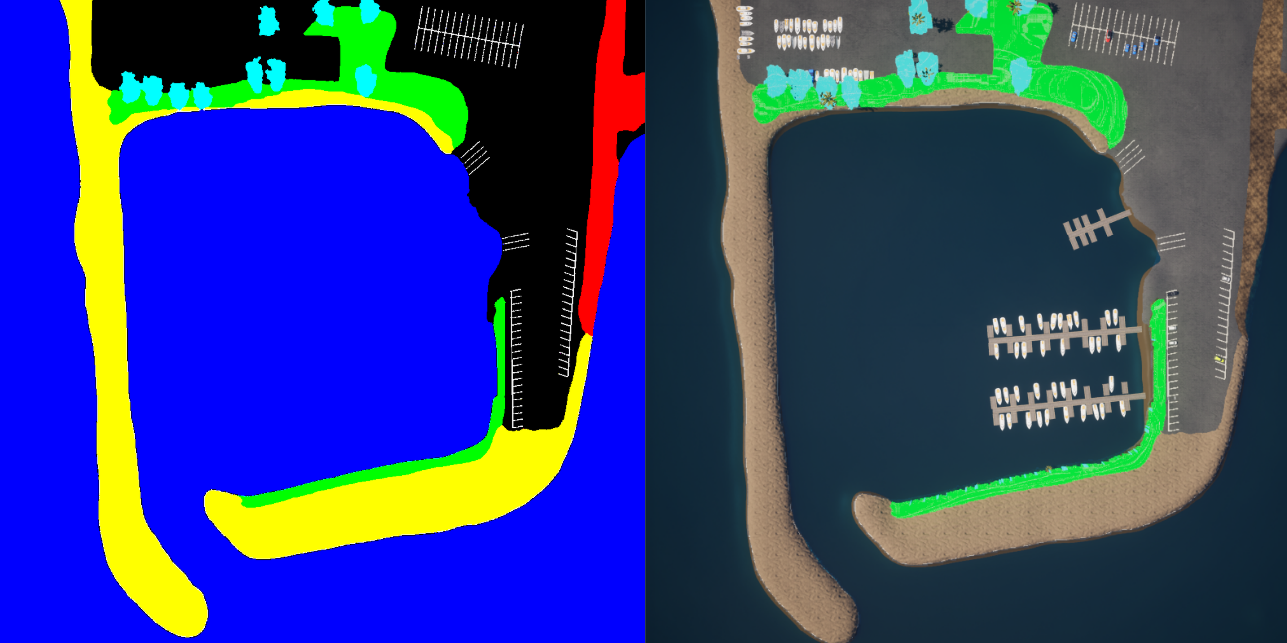}
    \caption{Example biome map and corresponding bounding boxes for procedural vegetation placement highlighted in UE.}
    \label{fig:vegetation}
\end{figure}

\subsection{Asset Library Generation}

Levels are generated with user provided assets, and as such users can utilize the UE marketplace to find assets that meet their exact needs. First, UE is used to automate the generation of rendered images of each asset. These rendered images are captured with three different lighting conditions and two different background colors compiled into a single library to improve chances of matching. Those images are then inputted into DINOv3~\cite{simeoni2025dinov3}, a vision transformer model, to create a database of asset embedding.

\begin{figure*}[t]
    \centering

    \begin{subfigure}[t]{0.24\textwidth}
        \centering
        \includegraphics[width=\linewidth]{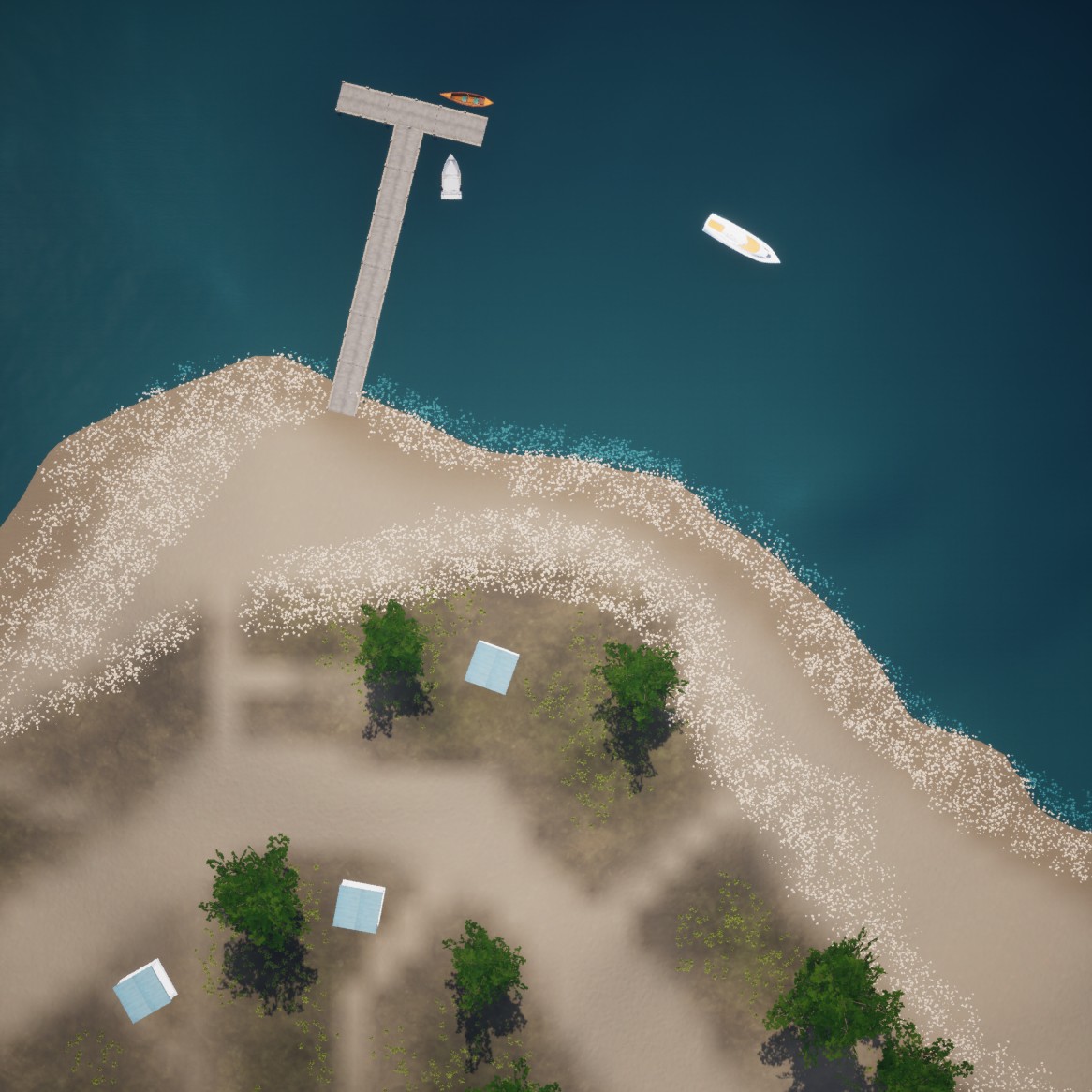}
        \caption*{\textbf{Level 1} -- Ground Truth}
    \end{subfigure}
    \begin{subfigure}[t]{0.24\textwidth}
        \centering
        \includegraphics[width=\linewidth]{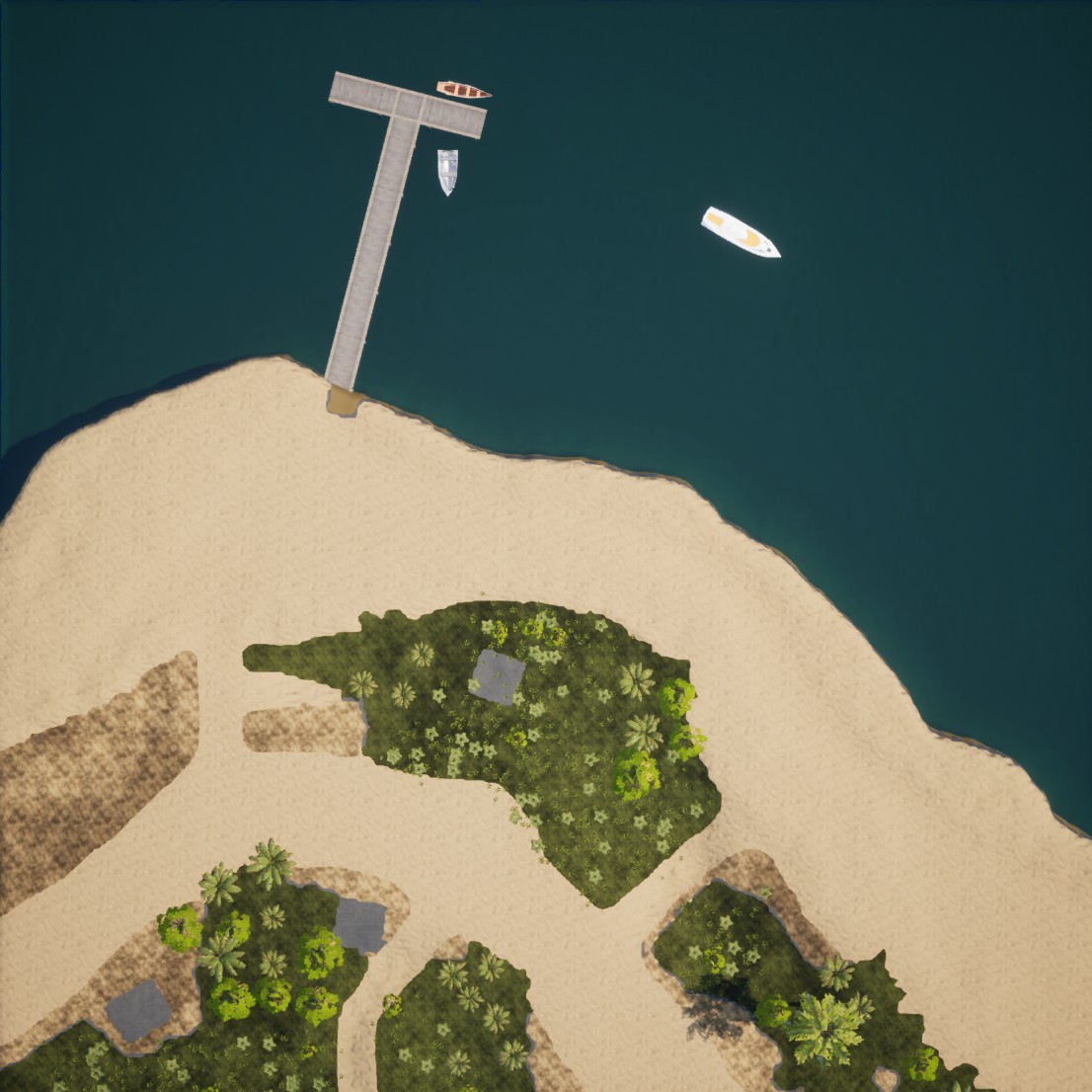}
        \caption*{\textbf{Level 1} -- Replicated}
    \end{subfigure}
    \begin{subfigure}[t]{0.24\textwidth}
        \centering
        \includegraphics[width=\linewidth]{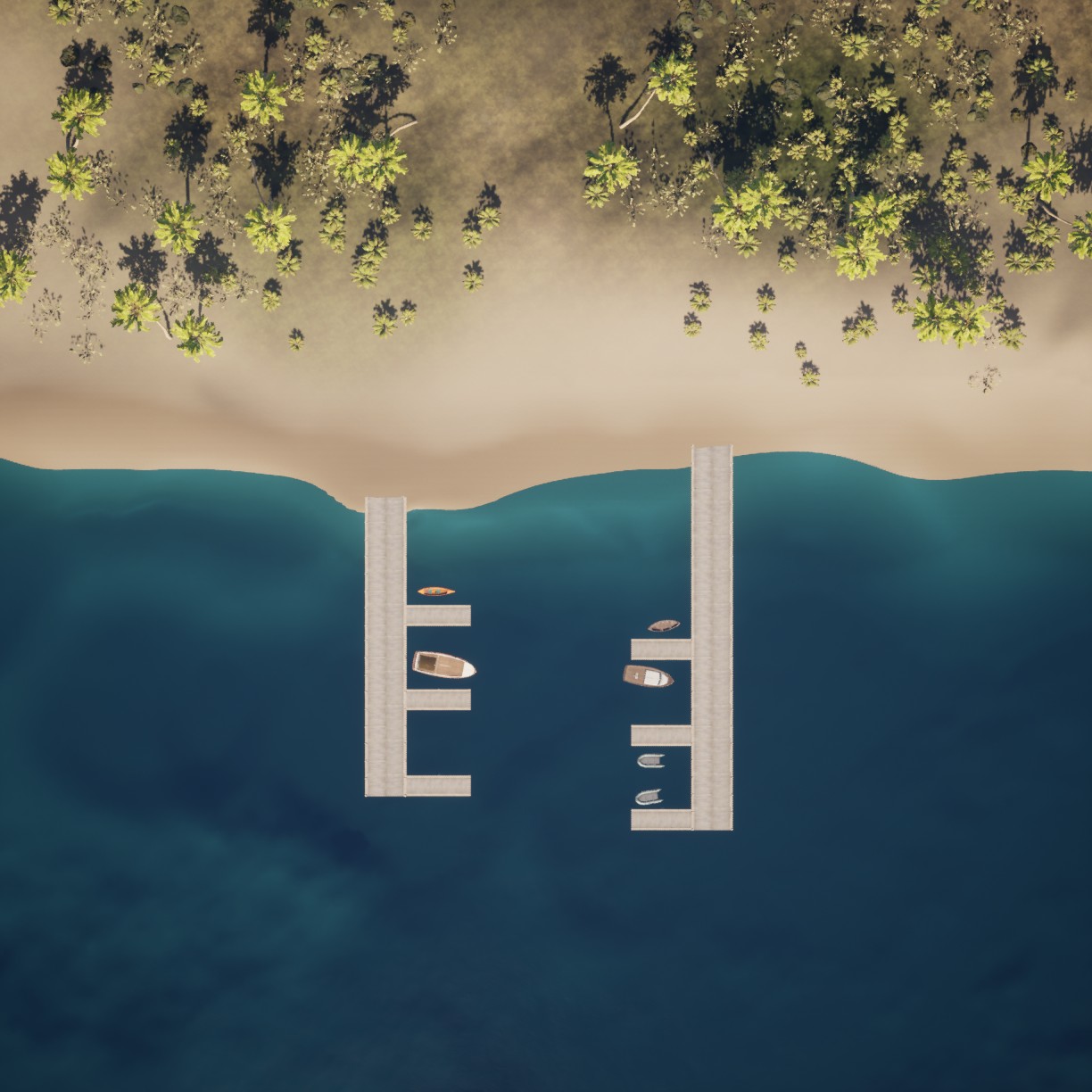}
        \caption*{\textbf{Level 2} -- Ground Truth}
    \end{subfigure}
    \begin{subfigure}[t]{0.24\textwidth}
        \centering
        \includegraphics[width=\linewidth]{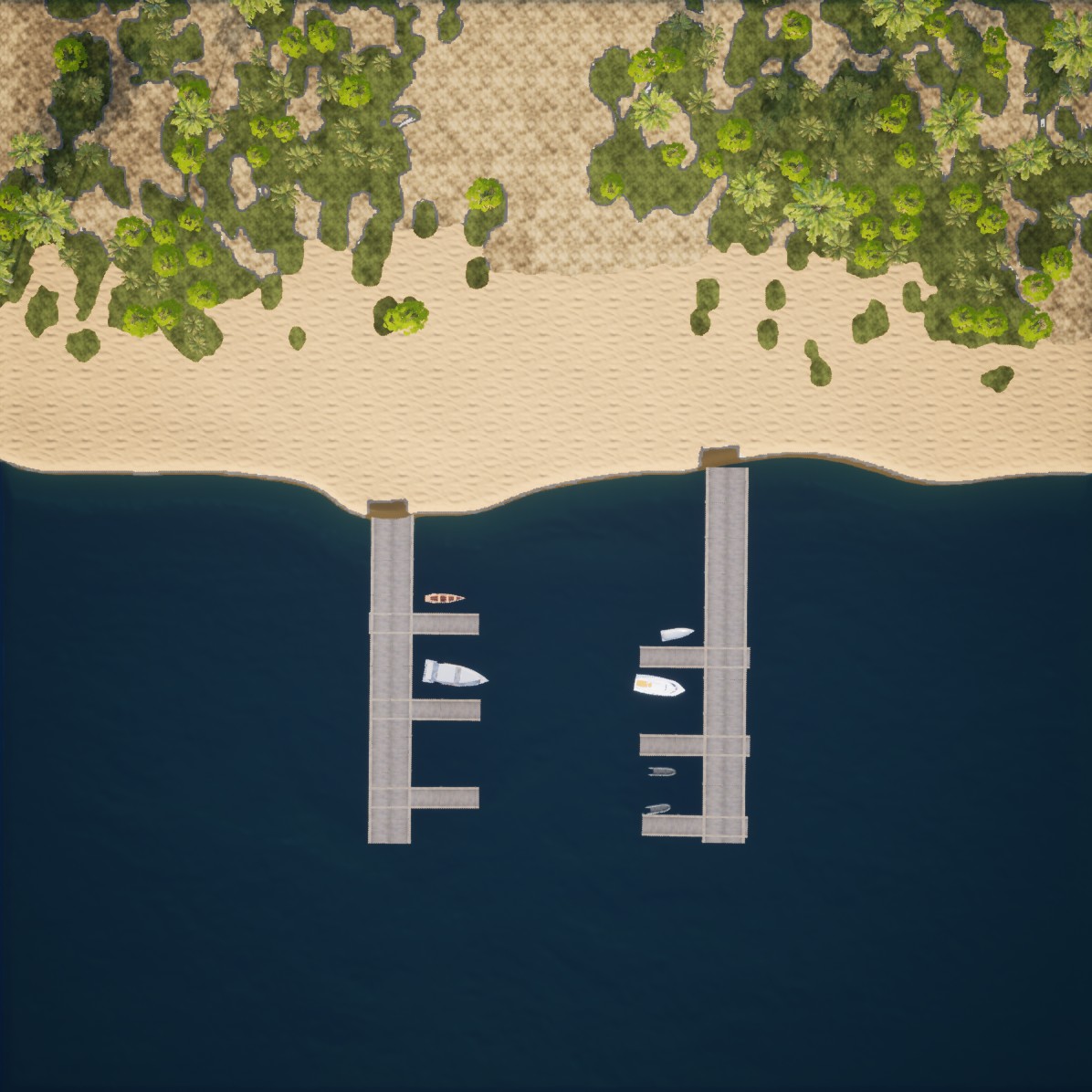}
        \caption*{\textbf{Level 2} -- Replicated}
    \end{subfigure}

    \begin{subfigure}[t]{0.24\textwidth}
        \centering
        \includegraphics[width=\linewidth]{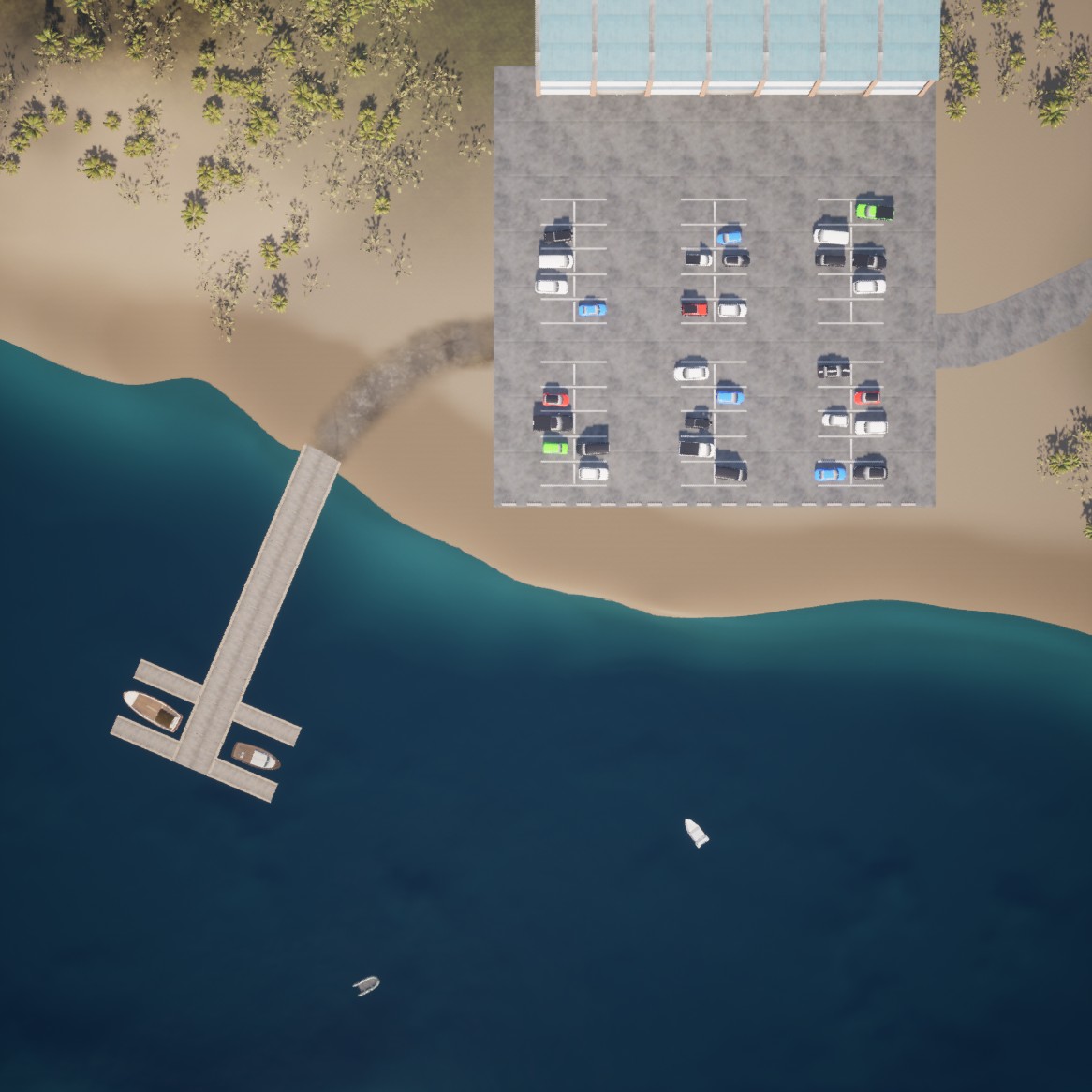}
        \caption*{\textbf{Level 3} -- Ground Truth}
    \end{subfigure}
    \begin{subfigure}[t]{0.24\textwidth}
        \centering
        \includegraphics[width=\linewidth]{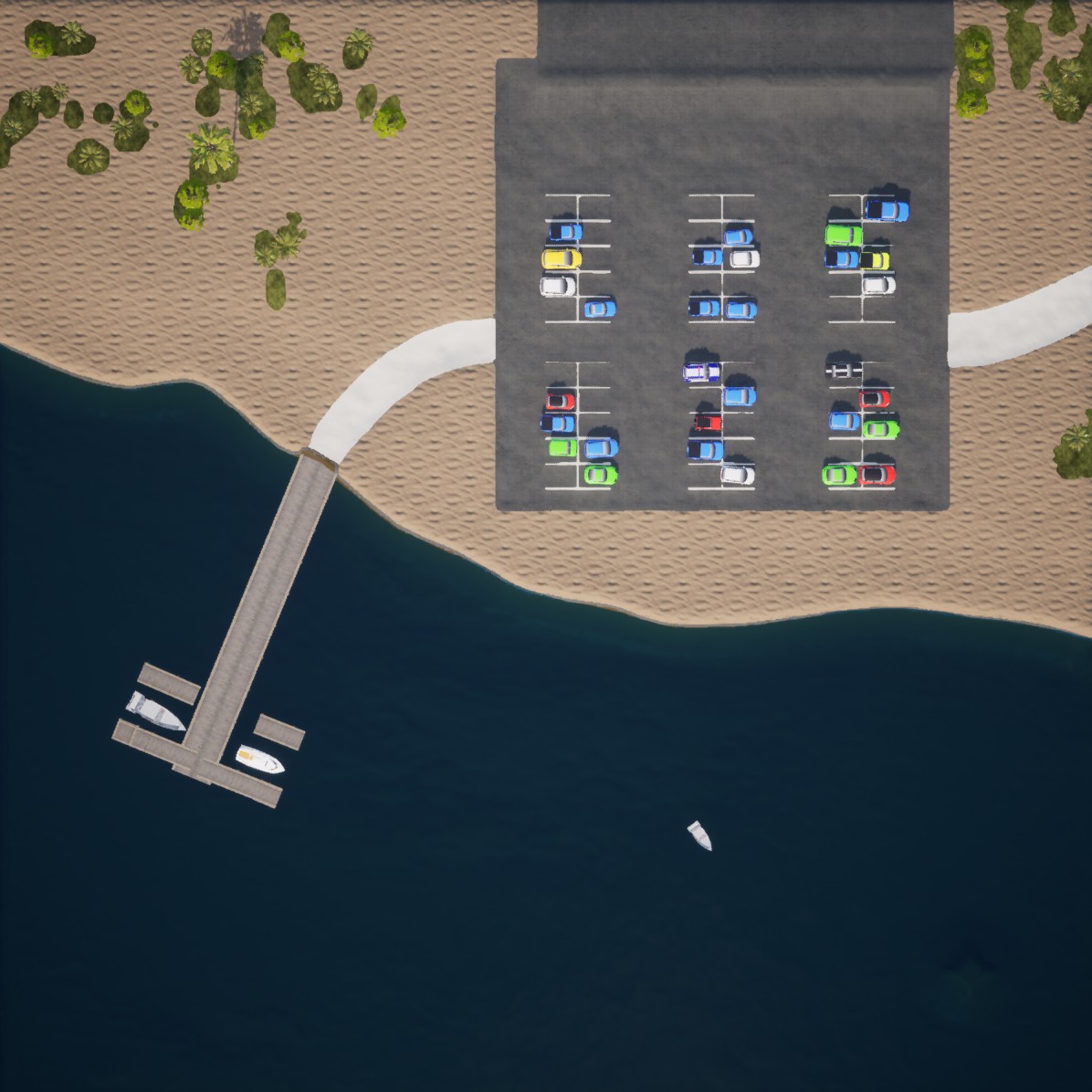}
        \caption*{\textbf{Level 3} -- Replicated}
    \end{subfigure}
    \begin{subfigure}[t]{0.24\textwidth}
        \centering
        \includegraphics[width=\linewidth]{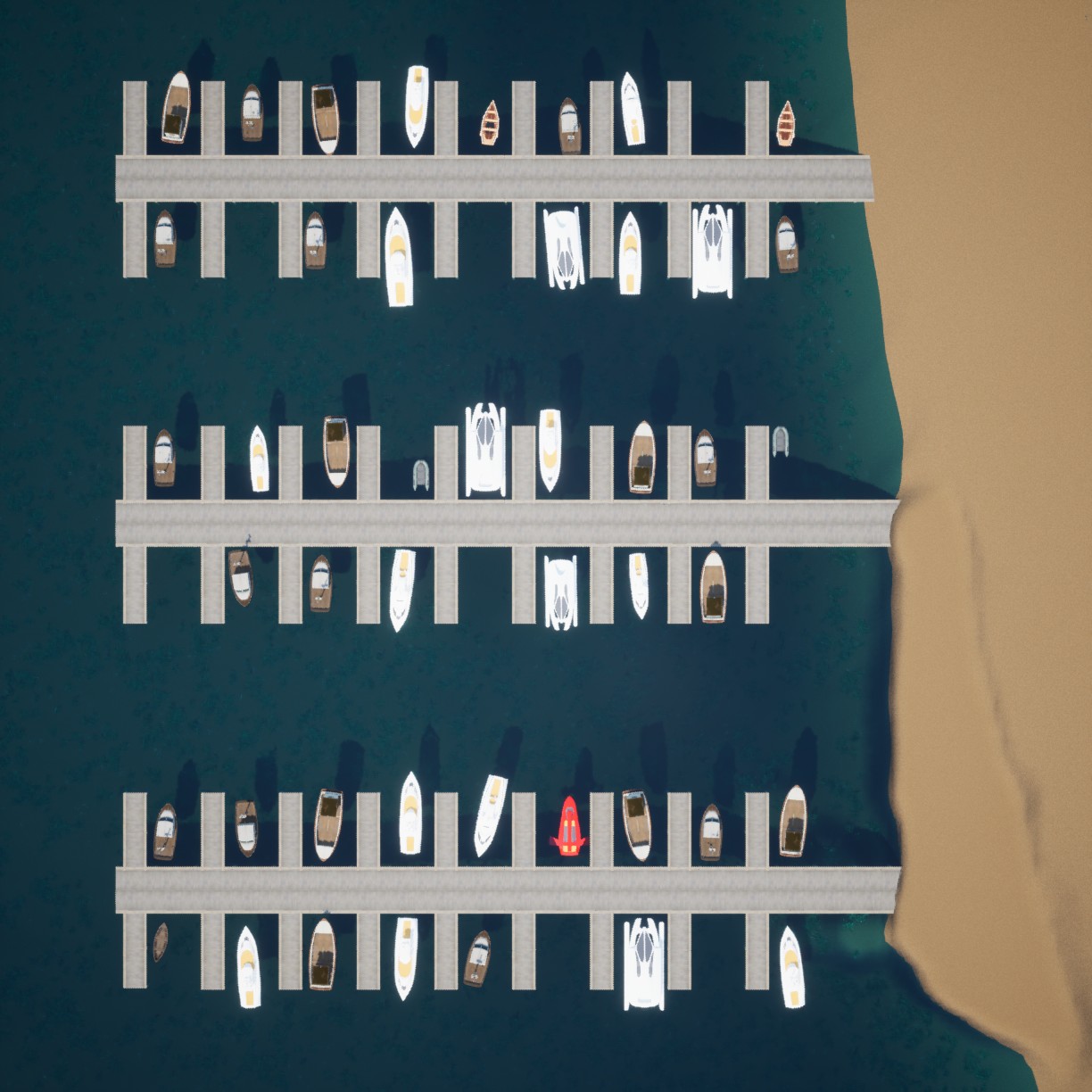}
        \caption*{\textbf{Level 4} -- Ground Truth}
    \end{subfigure}
    \begin{subfigure}[t]{0.24\textwidth}
        \centering
        \includegraphics[width=\linewidth]{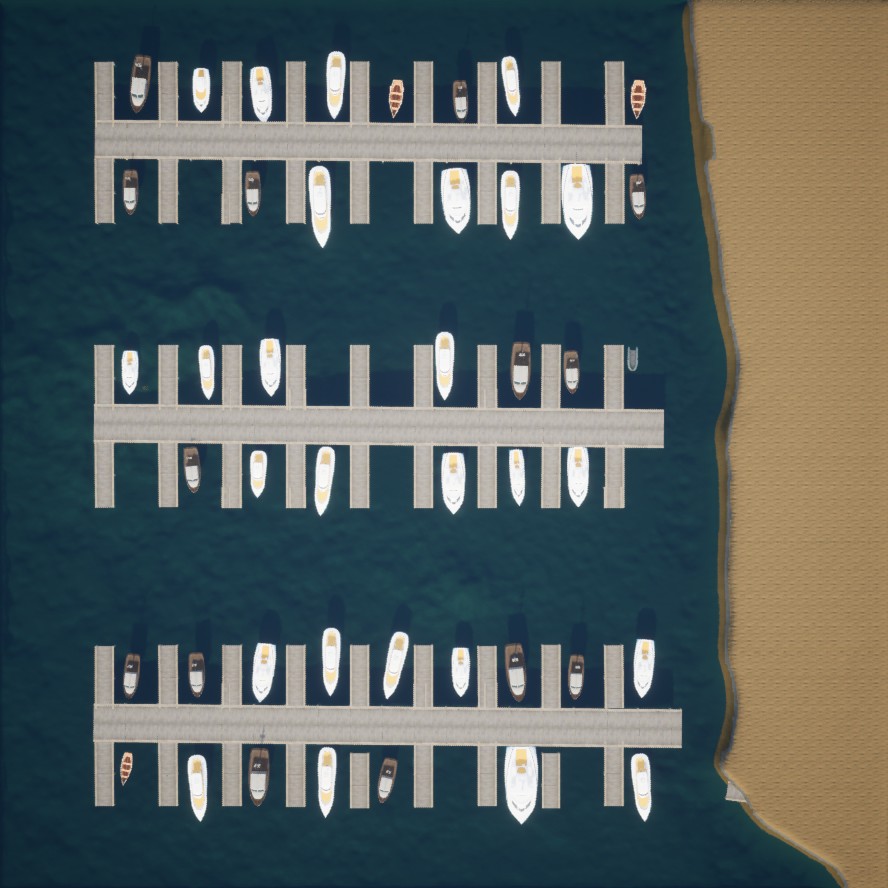}
        \caption*{\textbf{Level 4} -- Replicated}
    \end{subfigure}

    \makebox[\textwidth][c]{%
        \begin{subfigure}[t]{0.24\textwidth}
            \centering
            \includegraphics[width=\linewidth]{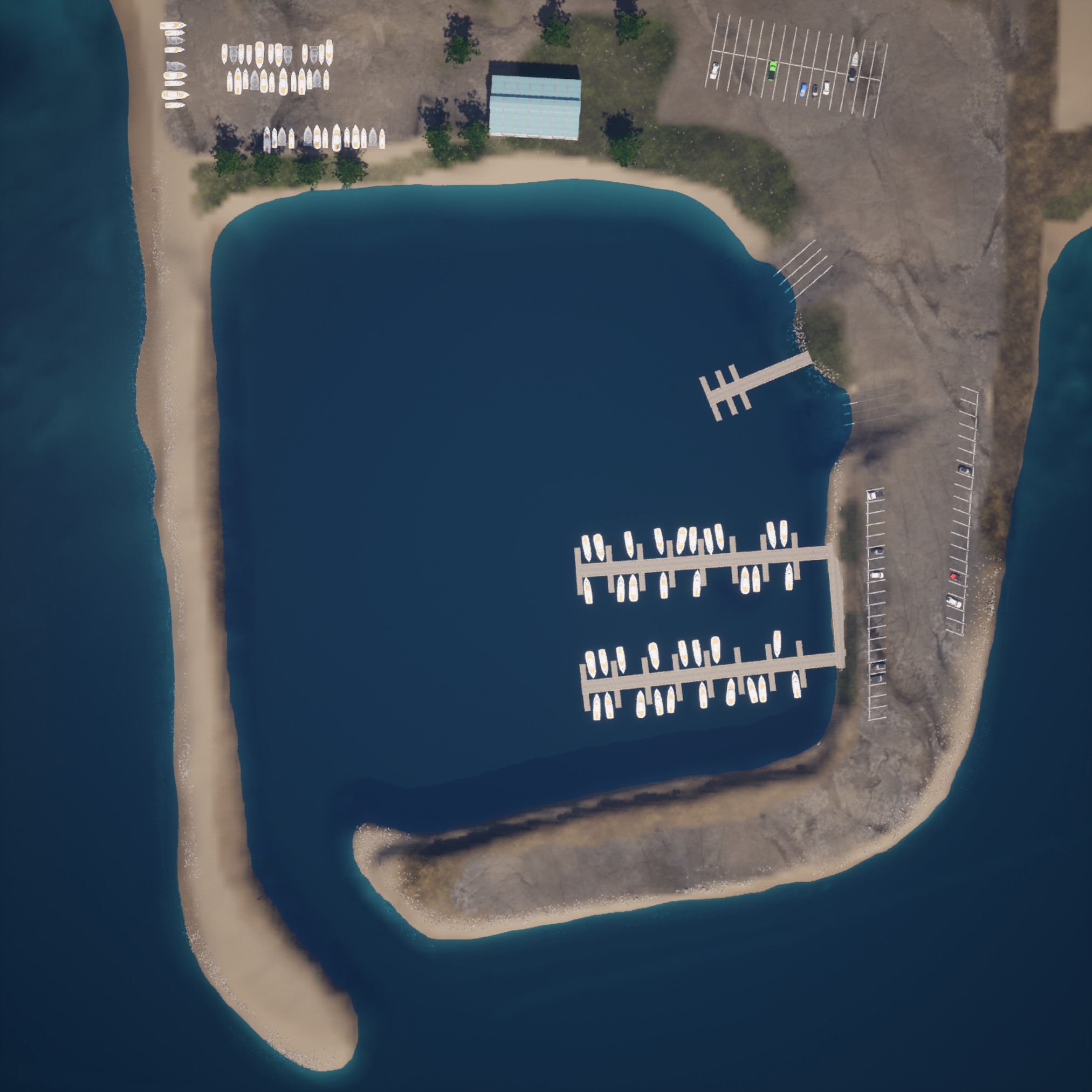}
            \caption*{\textbf{Level 5} -- Ground Truth}
        \end{subfigure}
        \begin{subfigure}[t]{0.24\textwidth}
            \centering
            \includegraphics[width=\linewidth]{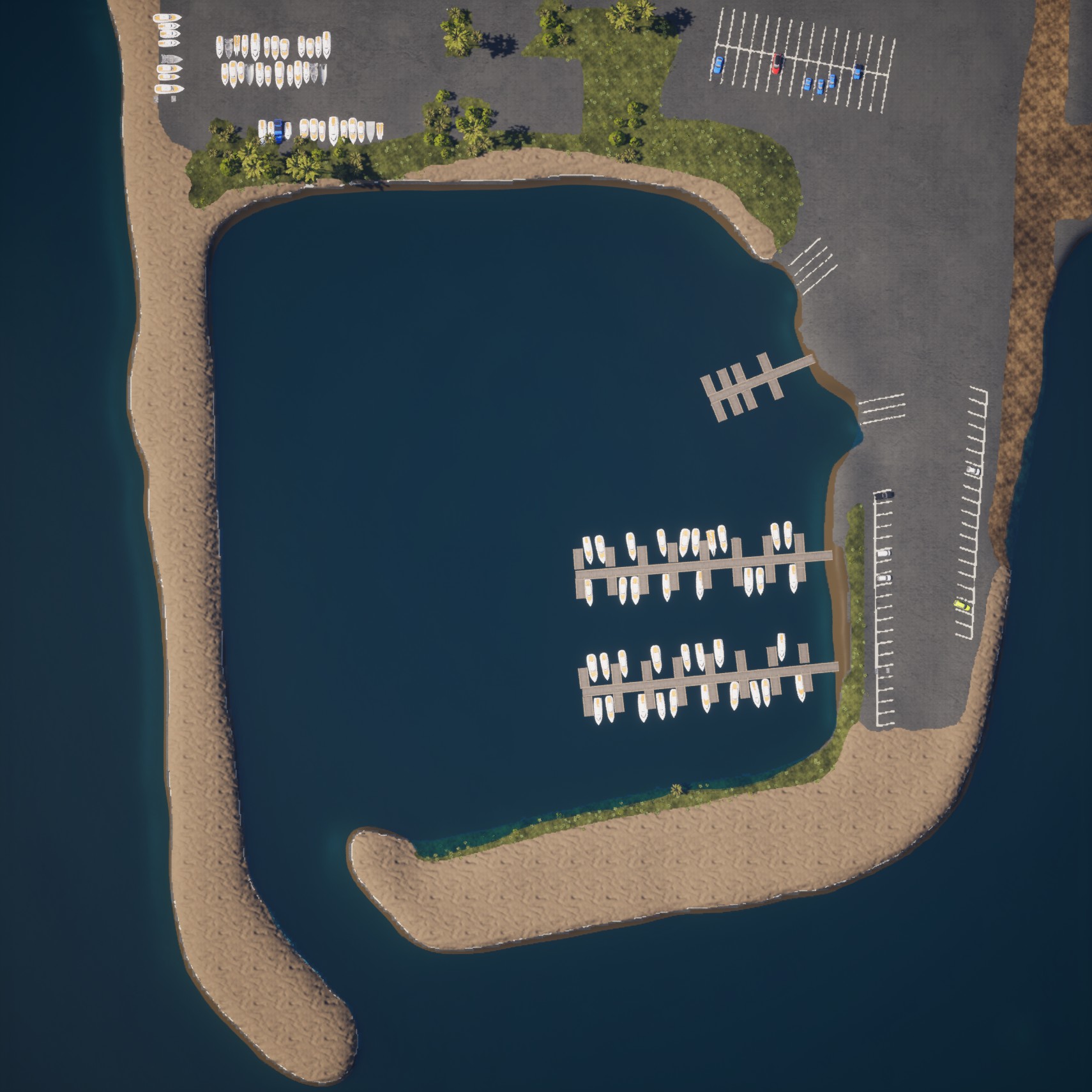}
            \caption*{\textbf{Level 5} -- Replicated}
        \end{subfigure}
    }

    \caption{Qualitative comparison between the ground-truth simulated scenes
    and their corresponding replicated scenes.}
    \label{fig:qualitative_comparison}
\end{figure*}

\subsection{Asset Selection and Placement System}

Our asset selection and placement system manages both manmade objects such as boats and piers, and natural objects such as vegetation, trees, and rocks. 

For manmade objects, YOLO~\cite{yolo26} is used on the user-provided aerial image to find candidate locations for asset placement. The associated object oriented bounding box image clips for these detected objects are used as input for DINOv3 to generate embedding vectors for the objects. Using Facebook AI Similarity Search (FAISS)~\cite{johnson2019billion}, each object embedding is compared to the asset library embedding database generated perviously and a random close match is chosen based on cosine similarity. Each asset's location in the level is converted from pixel location to UE location using the dimensions of the original image as the conversion rate. Asset scales and rotations are also determined from the YOLO generated bounding boxes. 

For piers, two more YOLO passes are performed on the cleaned image generated in Section \ref{sec:Semantic}: one for detecting the main walkways, and one for detecting the smaller pier parts. For placement, the bounding box is used to define the bounds of the pier, and a user selected pier asset is spawned in a chain to fit the bounds. 

Natural object placement is handled through the UE PCG system. The semantic segmentation map created in step \ref{sec:Semantic} is used as a biome map for the PCG system, as demonstrated in Fig.~\ref{fig:vegetation}. The biome map defines rules for each terrain type with various predefined filters to determine location, density, and type of vegetation. An elevation filter prevents trees from being spawned on slopes but allows rocks and grass to be spawned. A water level filter is used to differentiate between the placement of underwater and land vegetation.

\section{Evaluation}

\begin{table*}[t]
\centering
\begin{tabular}{lcccccccc}
\toprule
\textbf{Level} &
\textbf{Objects} &
\textbf{Resolution} &
\textbf{m/pixel} &
\makecell{\textbf{XY} \\ \textbf{RMSE (m)}} &
\makecell{\textbf{Rotation} \\ \textbf{RMSE ($^\circ$)}} &
\makecell{\textbf{Orientation} \\ \textbf{RMSE ($^\circ$)}} &
\makecell{\textbf{Precision}} &
\makecell{\textbf{Recall}} \\
\midrule

\multirow{2}{*}{Level 1}
& \multirow{2}{*}{3}
& Low & 0.188 & 0.249 & 146.9 & 0.19 & 1 & 1 \\
&
& High & 0.094 & 0.145 & 146.9 & 0.22 & 1 & 1 \\
\midrule

\multirow{2}{*}{Level 2}
& \multirow{2}{*}{6}
& Low & 0.258 & 0.303 & 126.4 & 1.52 & 0.5 & 1 \\
&
& High & 0.129 & 0.220 & 126.9 & 0.73 & 1 & 1 \\
\midrule

\multirow{2}{*}{Level 3}
& \multirow{2}{*}{34}
& Low & 0.273 & 0.354 & 115.7 & 2.03 & 0.839 & 0.912 \\
&
& High & 0.136 & 0.327 & 120.4 & 1.71 & 1 & 0.971 \\
\midrule

\multirow{2}{*}{Level 4}
& \multirow{2}{*}{46}
& Low & 0.228 & 0.438 & 158.7 & 4.19 & 1 & 0.935 \\
&
& High & 0.114 & 0.460 & 159.2 & 4.07 & 1 & 0.957 \\
\midrule

\multirow{2}{*}{Level 5}
& \multirow{2}{*}{96}
& Low & 0.254 & 0.900 & 126.6 & 1.52 & 0.936 & 0.979 \\
&
& High & 0.170 & 0.550 & 126.0 & 1.20 & 0.957 & 0.965 \\

\midrule
\multicolumn{9}{c}{\textbf{Average Performance Across Levels}} \\
\midrule

\multicolumn{2}{l}{Low Resolution}
& -- & -- & 0.448 & 134.9 & 1.89 & 0.855 & 0.965 \\

\multicolumn{2}{l}{High Resolution}
& -- & -- & 0.340 & 135.9 & 1.58 & 0.991 & 0.978  \\

\midrule
\multicolumn{9}{c}{\textbf{Results for Real World Imagery}} \\
\midrule
Harbor & 253 & High & 0.042 & 0.329 & - & 2.45 & 0.972 & 0.838 \\

\bottomrule
\end{tabular}
\caption{Results for comparison between ground truth and replicated environments.}
\label{tab:resolution_results}
\end{table*}

\subsection{Synthetic Environment Regeneration Experiments}
For evaluation, we manually created five simulated environments in UE and then captured overhead imagery of these levels to serve as ground truth. These simulated environments include two simple pier scenes, a pier with a parking lot, a section of HoloOcean's PierHarbor level, and a lake harbor. For each simulated environment, a UE camera was used to capture aerial imagery at two levels of pixel resolution. Fig.~\ref{fig:qualitative_comparison} has image comparisons between the input and pipeline generated output environments. The only modification made was fine tuning the color multipliers for the landscape texture categories to better match the input images.  

Comparison between pixel resolutions on each environment are recorded in Table~\ref{tab:resolution_results}. We recorded the x and y locations and yaw rotations for each asset in both the ground truth and replicated environments. We then calculated xy location RMSE and rotation RMSE by comparing these values for each asset. Along with rotation RMSE, orientation RMSE was calculated by wrapping yaw rotation measurements at 180 degrees back to zero. This allowed us to compare whether or not assets were aligned on the same axis. For this assessment, precision indicates the ratio of correctly matched objects for each scene, and recall indicates the ratio of objects found.

We found that our xy location RMSE for asset placement averaged less than 0.5 meters. Although rotation RMSE was high, orientation RMSE was less than 2 degrees. This is because asset rotations are determined purely from the YOLO object oriented bounding box, and we do not account for the direction an object is facing. To accommodate for this, 40\% of the time the pipeline adds 180 degrees to flip the direction of an asset. This flip was not implemented for these results to get unaltered rotational errors. 

Although both low and high resolution inputs had comparable asset location error, the low resolution inputs had many more object mismatches. Increasing the resolution of the input improved precision from 0.859 to 0.992, whereas recall only increased from 0.965 to 0.978. Input resolution primarily affects object matching rather than object detection. 

For pier generation, YOLO sometimes failed to grab the entire bounding box for the pier. This can best be seen in the Level 3 and Level 4 generated results in Fig.~\ref{fig:qualitative_comparison} as some of the pier parts do not connect to each other or the land. 

The Gemini-created semantic segmentation maps mostly identified regions for vegetation, although it didn't allow for fine details. As seen in the Level 1 generated results, it determined that the entire dirt area was grass and as a result spawned grass over the entire area. Although the segmentation map determines areas for different vegetation, users can always fine tune the density of these objects to fit their needs.

\begin{figure*}[t]
    \centering

    \begin{subfigure}[t]{0.49\textwidth}
        \centering
        \includegraphics[width=\linewidth]{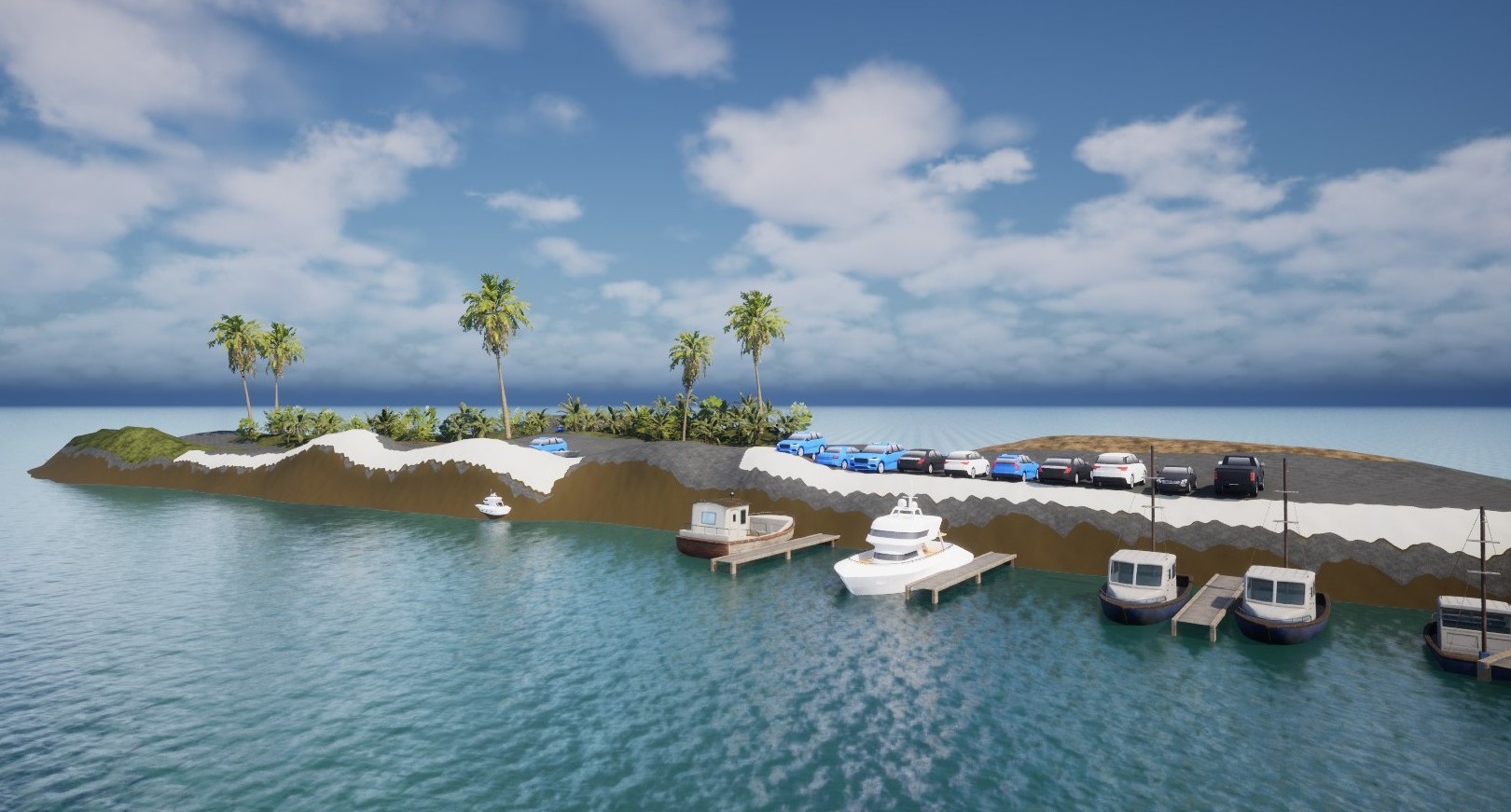}
        \caption{Side view of generated Harbor scene.}
    \end{subfigure}
    \hfill
    \begin{subfigure}[t]{0.49\textwidth}
        \centering
        \includegraphics[width=\linewidth]{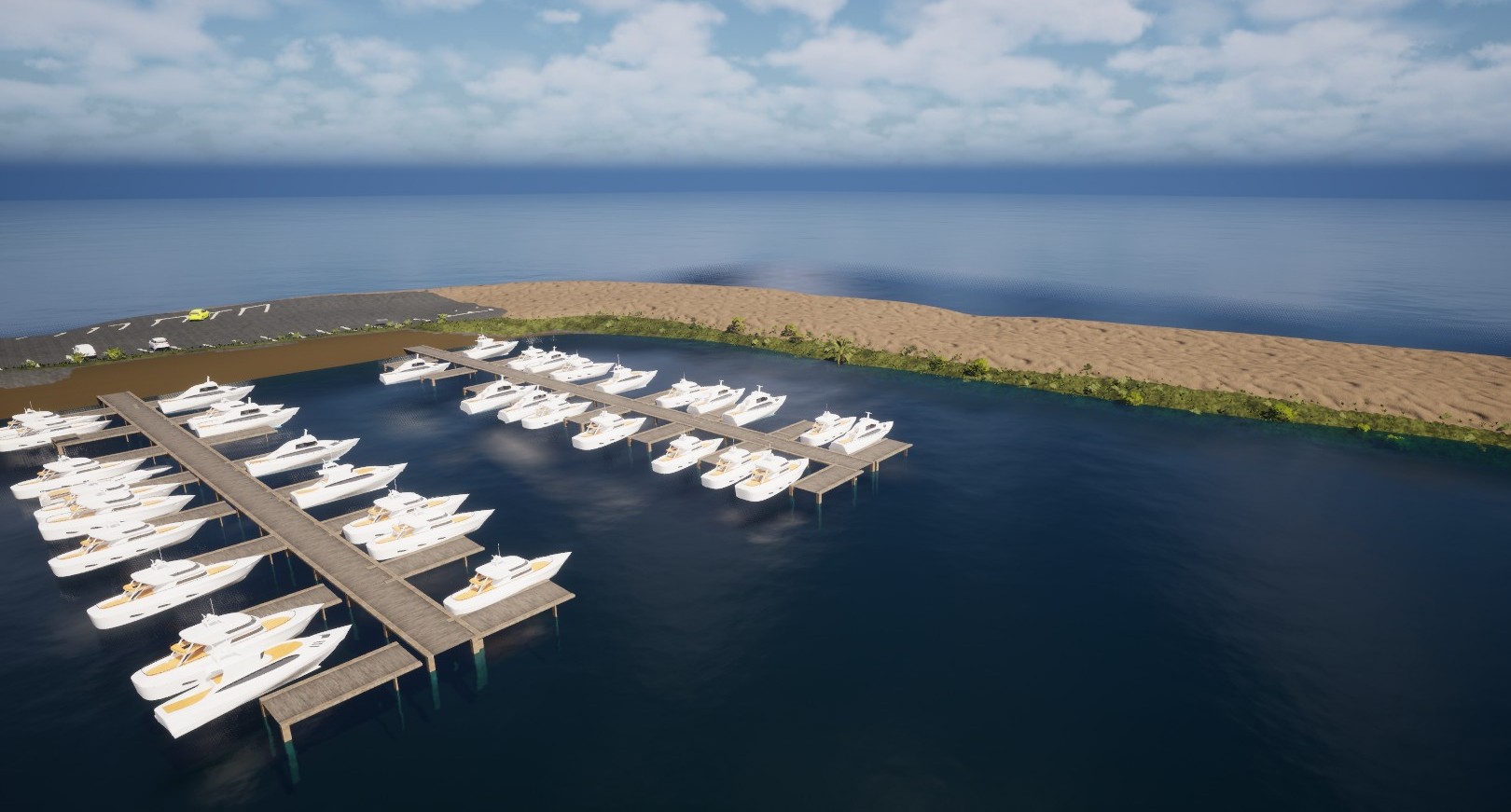}
        \caption{Side view of generated Level 5.}
    \end{subfigure}

    \caption{Comparison between two different height map results from Seabed-Net. The model performed poorly on our real world harbor example, but created a closer match for our Level 5 example.}
    \label{fig:side_view_comparison}
\end{figure*}

Seabed-Net was trained for bathymetry, so the height maps it generated were not ideal; however, it provided a fair starting point that could be further edited by a user. The model could identify the land and water portions of overhead imagery, but created stark differences between the two, with no ability for gradual changes in terrain, as demonstrated in Fig.~\ref{fig:side_view_comparison}. While Seabed-net provides non-ideal results, we note that the proposed framework is agnostic to the bathymetry and DEM generation model. We are actively working on an improved terrain and bathymetry generation pipeline that will be presented in future work.

The results were generated using a NVIDIA GeForce RTX 5070 GPU. The initial setup time to create the asset embedding database can vary depending on the number of user provided assets. Our database contained 159 assets and the system required approximately 44 minutes and 45.6 seconds up front to generate the embeddings and asset database. This is because we calculated 5 different embeddings per asset to match different lighting conditions. Users could choose to only use one embedding and decrease the initial setup time. Once these embeddings are generated, this step does not need to be repeated as we were able to use the same database across all generated results. YOLO took on average 0.31 seconds per slice, with the number of slices varying for different resolutions. DINOv3 took on average 3.27 seconds per asset match, and Seabed-Net took 6.1 seconds to generate a depth map. Level 1 was the smallest level at only 3 assets, and in total took 20.13 seconds to generate. Level 5 with 96 assets took approximately 5 minutes and 39.2 seconds to generate.

\subsection{Real World Imagery Test}
As a real world example, we used aerial drone imagery of Hale‘iwa Boat Harbor taken by the FROST Lab. The drone imagery was hand labeled to act as a ground truth for the pipeline generated level and the results can be found in Table~\ref{tab:resolution_results}. Rotational RMSE was not calculated for the real world example because the hand labeling did not account for object direction. Although the precision and recall are slightly lower than the simulated results, the visuals of the generated level match very closely to the real world input as depicted in Fig.~\ref{fig:real_harbor_comparison}. The texture placement matches very closely to the real world image, and the vegetation spawns in proper locations. Although the majority of the objects were detected, the pipeline struggled to detect objects that were partially obscured or too close to other objects. There were a handful of instances where the correct class was identified, but the wrong type of object was selected; for example, a police car was spawned in place of a normal car, and a yacht was spawned where a small raft should be.

\section{Conclusion}

We developed a pipeline for coastal environment generation for HoloOcean. The pipeline enables landscape generation through the integration of a model for height maps, while also utilizing Google Gemini to create semantic segmentation maps for texturing and vegetation placement. Our pipeline also enables asset matching through YOLO to detect objects and then selects similar assets with DINOv3. 

Plans for future work involve developing a custom DEM and bathymetry model that will replace Seabed-Net. This model will be trained on pairs of RGB imagery and DEM data. 

Although the generation time scales linearly with the number of detected assets, the final time is on the offer of seconds rather than the days required to manually create a complicated coastal level. 

Future work will continue to improve upon this pipeline and enable easier user customization of generated levels. 

\bibliographystyle{IEEEtran}
\IEEEtriggeratref{19}
\bibliography{ref}

@article{simeoni2025dinov3,
  title={{DINOv3}},
  author={Sim{\'e}oni, Oriane and others},
  journal={arXiv:2508.10104},
  year={2025},
}

@INPROCEEDINGS{DEM_w_deeplearning,
  author={Madani, Alif Ilham and Kuswati, Riska A. and Lechner, Alex M. and Saputra, Muhamad Risqi U.},
  booktitle={IEEE International Geoscience and Remote Sensing Symposium (IGARSS)}, 
  title={Digital Elevation Model Estimation from RGB Satellite Imagery Using Generative Deep Learning}, 
  year={2025},
  pages = "6291--6295",
  address = "USA",
  doi={10.1109/IGARSS55030.2025.11243010}}

@article{DEM_lunar,
author = {Chen, Tianhao and Wang, Yexin and Nan, Jing and Chenxu, Zhao and Wang, Biao and Xie, Bin and Liu, Wai Chung and Di, Kaichang and Liu, Bin and Chen, Shaohua},
year = {2025},
month = sep,
pages = {3097},
title = {A Generative Adversarial Network for Pixel-Scale Lunar DEM Generation from Single High-Resolution Image and Low-Resolution DEM Based on Terrain Self-Similarity Constraint},
volume = {17},
journal = {Remote Sensing},
doi = {10.3390/rs17173097}
}

@article{pixels2peaks,
author = {Jain, Aryamaan and Gain, James and Cordonnier, Guillaume},
title = {Pixels2Peaks: Converting Terrain Images to Heightmaps},
year = {2026},
publisher = {Association for Computing Machinery},
address = {New York, NY, USA},
volume = {45},
number = {4},
issn = {0730-0301},
doi = {10.1145/3811288},
journal = {ACM Trans. Graph.},
month = jul,
}

@INPROCEEDINGS{magicbathynet,
  author={Agrafiotis, Panagiotis and Janowski, {\L}ukasz and Skarlatos, Dimitrios and Demir, Beg{\"u}m},
  booktitle={IEEE International Geoscience and Remote Sensing Symposium (IGARSS)}, 
  title={MAGICBATHYNET: A Multimodal Remote Sensing Dataset for Bathymetry Prediction and Pixel-Based Classification in Shallow Waters}, 
  year={2024},
  month = jul,
  pages = {249-253},
  doi={10.1109/IGARSS53475.2024.10641355}}

@article{seabednet,
title = {Seabed-Net: A multi-task network for joint bathymetry estimation and seabed classification from remote sensing imagery in shallow waters},
journal = {Journal of Photogrammetry and Remote Sensing (ISPRS)},
volume = {231},
pages = {414-430},
year = {2026},
issn = {0924-2716},
doi = {https://doi.org/10.1016/j.isprsjprs.2025.11.007},
author = {Panagiotis Agrafiotis and Begüm Demir},
}

@article{johnson2019billion,
  title={Billion-scale similarity search with {GPUs}},
  author={Johnson, Jeff and Douze, Matthijs and J{\'e}gou, Herv{\'e}},
  journal={IEEE Transactions on Big Data},
  volume={7},
  number={3},
  pages={535--547},
  year={2019},
  publisher={IEEE}
}

@article{zhou2024scenexproceduralcontrollablelargescale,
      title={Scene{X}: Procedural Controllable Large-scale Scene Generation}, 
      author={Mengqi Zhou and Yuxi Wang and Jun Hou and Shougao Zhang and Yiwei Li and Chuanchen Luo and Junran Peng and Zhaoxiang Zhang},
      year={2024},
      journal={arXiv:2403.15698},
      archivePrefix={arXiv},
      primaryClass={cs.CV},
}

@inproceedings{autoue,
    title = "{A}uto{UE}: Automated Generation of 3{D} Games in Unreal Engine via Multi-Agent Systems",
    author = "Yin, Lei  and
      Cheng, Wentao  and
      Qin, Zhida  and
      Huang, Tianyu  and
      Li, Yidong  and
      Ding, Gangyi",
    booktitle = "Findings of the {A}ssociation for {C}omputational {L}inguistics: {ACL}",
    month = jul,
    year = "2026",
    address = "San Diego, California, USA",
    doi = "10.18653/v1/2026.findings-acl.111",
    ISBN = "979-8-89176-395-1",
}

@inproceedings{infinigen2023infinite,
  title={Infinite Photorealistic Worlds Using Procedural Generation},
  author={Raistrick, Alexander and Lipson, Lahav and Ma, Zeyu and Mei, Lingjie and Wang, Mingzhe and Zuo, Yiming and Kayan, Karhan and Wen, Hongyu and Han, Beining and Wang, Yihan and Newell, Alejandro and Law, Hei and Goyal, Ankit and Yang, Kaiyu and Deng, Jia},
  booktitle={Proceedings of the IEEE/CVF Conference on Computer Vision and Pattern Recognition (CVPR)},
  year={2023},
  pages={12630--12641}
}

@inproceedings{songtang-etal-2025-unrealllm,
    title = "{U}nreal{LLM}: Towards Highly Controllable and Interactable 3{D} Scene Generation by {LLM}-powered Procedural Content Generation",
    author = "Tang, Song  and
      Zhao, Kaiyong  and
      Wang, Lei  and
      Li, Yuliang  and
      Liu, Xuebo  and
      Zou, Junyi  and
      Wang, Qiang  and
      Chu, Xiaowen",
    booktitle = "Findings of the Association for Computational Linguistics: ACL",
    month = jul,
    year = "2025",
    address = "Vienna, Austria",
    pages = "19417--19435",
    doi = "10.18653/v1/2025.findings-acl.994",
}

@inproceedings{potokarHoloOcean2022,
  title = {{{HoloOcean}}: {{An}} Underwater Robotics Simulator},
  booktitle = {{{IEEE International Conference}} on {{Robotics}} and {{Automation (ICRA)}}},
  author = {Potokar, E. and Ashford, S. and Kaess, M. and Mangelson, J.},
  year = {2022},
  address = {{Philadelphia, PA, USA}}

}

@inproceedings{song2025oceansim,
      title={{OceanSim}: A {GPU}-Accelerated Underwater Robot Perception Simulation Framework}, 
      author={Jingyu Song and Haoyu Ma and Onur Bagoren and Advaith V. Sethuraman and Yiting Zhang and Katherine A. Skinner},
      booktitle={IEEE/RSJ International Conference on Intelligent Robots and Systems (IROS)},
      year={2025},
}

@inproceedings{UNav-Sim,
  title={{UNav-Sim}: A Visually Realistic Underwater Robotics Simulator and Synthetic Data-generation Framework},
  author={Amer, Abdelhakim and {\'A}lvarez-Tu{\~n}{\'o}n, Olaya and U{\u{g}}urlu, Halil {\.I}brahim and Sejersen, Jonas Le Fevre and Brodskiy, Yury and Kayacan, Erdal},
  booktitle={21st IEEE International Conference on Advanced Robotics (ICAR)},
  year={2023},
  pages={570-576},
}

@STRING{C-OCEANS-Europe = {{IEEE OCEANS} Conference Proceedings}}

@inproceedings{stonefish,
    author = {Cie{\'s}lak, Patryk},
    booktitle = C-OCEANS-Europe,
    title = {{Stonefish: An Advanced Open-Source Simulation Tool Designed for Marine Robotics, With a ROS Interface}},
    year = {2019},
    pages={1-6},
    doi = {10.1109/OCEANSE.2019.8867434},
    address = {{Marseille, France}},
}

@misc{google_gemini,
  author       = {{Google}},
  title        = {Gemini},
  publisher    = {Google},
  url          = {https://gemini.google.com},
}

@INPROCEEDINGS{projectDAVE,
  author={Zhang, Mabel M. and Choi, Woen-Sug and Herman, Jessica and Davis, Duane and Vogt, Carson and McCarrin, Michael and Vijay, Yadunund and Dutia, Dharini and Lew, William and Peters, Steven and Bingham, Brian},
  booktitle={IEEE/OES Autonomous Underwater Vehicles Symposium (AUV)}, 
  title={{DAVE} Aquatic Virtual Environment: Toward a General Underwater Robotics Simulator}, 
  year={2022},
  doi={10.1109/AUV53081.2022.9965808}}

@misc{unrealengine,
  author       = {{Epic Games}},
  title        = {{Unreal Engine}},
  howpublished = {URL: https://www.unrealengine.com}
}

@misc{unrealPCG,
 author       = {{Epic Games}},
  title        = {{Unreal Engine Procedural Content Generation}},
  howpublished = {Available: https://dev.epicgames.com/documentation/unreal-engine/procedural-content-generation-overview}
}

@misc{cesium,
  author       = {{Cesium, GS Inc.}},
  title        = {{Cesium}},
  howpublished = {Available: https://cesium.com/ [software]},
  note         = {Version 2.25.0}
}

@article{yolo26,
      title={Ultralytics {YOLO}26: Unified Real-Time End-to-End Vision Models}, 
      author={Glenn Jocher and Jing Qiu and Mengyu Liu and Shuai Lyu and Fatih Cagatay Akyon and Muhammet Esat Kalfaoglu},
      year={2026},
      eprint={2606.03748},
      journal={arXiv:2606.03748},
      archivePrefix={arXiv},
      primaryClass={cs.CV},
}
\end{document}